# Efficient Active Algorithms for Hierarchical Clustering

Akshay Krishnamurthy                                                                      AKSHAYKR@CS.CMU.EDU
Sivaraman Balakrishnan                                                                    SBALAKRI@CS.CMU.EDU
Min Xu                                                                                         MINX@CS.CMU.EDU
Aarti Singh                                                                                   AARTI@CS.CMU.EDU
Carnegie Mellon University 5000 Forbes Avenue, Pittsburgh, PA 15213

## Abstract

Advances in sensing technologies and the growth of the internet have resulted in an explosion in the size of modern datasets, while storage and processing power continue to lag behind. This motivates the need for algorithms that are efficient, both in terms of the number of measurements needed and running time. To combat the challenges associated with large datasets, we propose a general framework for *active* hierarchical clustering that repeatedly runs an off-the-shelf clustering algorithm on small subsets of the data and comes with guarantees on performance, measurement complexity and runtime complexity. We instantiate this framework with a simple spectral clustering algorithm and provide concrete results on its performance, showing that, under some assumptions, this algorithm recovers all clusters of size $\Omega(\log n)$ using $O(n \log^2 n)$ similarities and runs in $O(n \log^3 n)$ time for a dataset of $n$ objects. Through extensive experimentation we also demonstrate that this framework is practically alluring.

## 1. Introduction

Clustering is a ubiquitous task in exploratory data analysis, data mining, and several application domains. In clustering, we assign each object to one or more groups so that objects in the same group are very similar while objects in different groups are dissimilar. In a hierarchical clustering, the groups have multiple resolutions, so that a large cluster may be recursively divided into smaller sub-clusters. There exist many effective algorithms for clustering, but as modern data sets get larger, the fact that these algorithms require *every* pairwise similarity between objects poses a serious measurement and/or computational burden and limits the practicality of these algorithms. It is therefore practically appealing to develop clustering algorithms that are effective on large scale problems from both a measurement and a computational perspective.

To achieve both measurement and computational improvements, we focus on reducing the number of similarity measurements required for clustering. This approach results in immediate reduction in measurement overhead in applications where similarities are observed directly, but it can also provide dramatic computational gains in applications where similarities between objects are computed via some kernel evaluated on observed object features. The case of internet topology inference is an example of the former, where covariance in the packet delays observed at nodes reflects the similarity between them. Obtaining these similarities requires injecting probe packets into the network and places a significant burden on network infrastructure. Phylogenetic inference and other biological sequence analyses are examples of the latter, where computationally intensive edit distances are often used. In both cases our algorithms are dramatically faster than many popular algorithms.

In this paper, we propose a novel framework for speeding up hierarchical clustering algorithms through *activization*–creating active versions of the algorithms where only a small number of informative similarities are measured. Our framework allows the user to specify various levels of activeness and we provide theoretical analysis that quantifies the resulting trade-off between measurement overhead and computation time on one hand, and statistical accuracy on the other.

As a detailed example, we apply our framework to spectral clustering. Spectral clustering is a very popular clustering technique that relies on the structure of the eigenvectors of the Laplacian of the similarity matrix. These algorithms have received considerable attention in recent years because of their empirical success, but they suffer from the fact that they require all $n(n-1)/2$ similarities between the $n$ objects to be clustered and must compute a spectral decomposition, which on large datasets can be computationally prohibitive. Our active algorithm avoids this limitation by subsampling few objects in each round and only computing eigenvectors of very small sub-matrices. By appealing to previous statistical guarantees (Balakrishnan et al., 2011), we can show that this algorithm has desirable theoretical properties, both in terms of statistical and computational performance.

## 2. Related Work

There is a large body of work on hierarchical and partitional clustering algorithms, many coming with various theoretical guarantees, but only few algorithms attempt to minimize the number of pairwise similarities used (Eriksson et al., 2011; Balcan & Gupta, 2010; Shamir & Tishby, 2011). Along this line, the work of Eriksson et. al. (2011) and Shamir and Tishby (2011) is closest in flavor to ours.

Eriksson et. al. (2011) develop an active algorithm for hierarchical clustering and analyze the correctness and measurement complexity of this algorithm under noise model where a small fraction of the similarities are inconsistent with the hierarchy. They show that for a constant fraction of inconsistent similarities, their algorithm can recover hierarchical clusters up to size $\Omega(\log n)$ using $O(n \log^2 n)$ similarities. Our analysis for ACTIVESPECTRAL yields similar results in terms of noise tolerance, measurement complexity, and resolution, but in the context of i.i.d. subgaussian noise rather than inconsistencies. Our algorithm is also computationally more efficient.

Another approach to minimizing the number of similarities used is via perturbation theory, which suggests that randomly sampling the entries of a similarity matrix preserves many of its important properties, such as its spectral norm (Achlioptas & McSherry, 2001). With this result, the Davis-Kahan theorem suggests that spectral clustering algorithms, which look at the eigenvectors of the Laplacian associated with the similarity matrix, can succeed in recovering the clusters. This intuition is formalized by Shamir and Tishby (2011) who analyze a binary spectral algorithm that randomly samples $b$ entries from the similarity matrix. Their results imply that as long as $b = \Omega(n \log^{3/2} n)$ their algorithm will find flat $k$-way clusters of size $\Omega(n)$ with high probability. Our work, translated to the flat clustering setting improves this guarantee; Theorem 2 implies that $O(n \log n)$ similarities are needed to recover the clustering. Furthermore, we can give guarantees on the size of smallest cluster $\Omega(\log n)$ that can be recovered in a hierarchy by *selectively* sampling similarities at each level.

Recently (Voevodski et al., 2012) proposed an active algorithm for flat $k$-way clustering that selects $O(k)$ landmarks and partitions the objects using distances to these landmarks. Theoretically, the authors guarantee approximate-recovery of clusters of size $\Omega(n)$ using $O(nk)$ pairwise distances. This idea of selecting landmarks bears strong resemblance to the first phase of our active clustering algorithm and also has connections to the Landmark MDS algorithm of de Silva and Tenenbaum (2002). These approaches are tied to specific algorithms, while our framework is much more general. Moreover, we guarantee exact cluster recovery (under mild assumptions) rather than approximate recovery, which translates into guarantees on hierarchical clustering.

A related direction is the body of work on efficient streaming and online algorithms for approximating the $k$-means and $k$-medians objectives (See for example (Charikar et al., 2003; Shindler et al., 2011)). As with (Voevodski et al., 2012), the guarantees for these algorithms do not immediately translate into an exact recovery guarantee, making it challenging to transform these approaches into hierarchical clustering algorithms. Moreover, the success of spectral clustering in practice suggests that an efficient spectral algorithm would also be very appealing. While there have been advances in this direction, the majority of these require the entire similarity matrix or graph to be known *a priori* (Frieze et al., 2004). Apart from (Shamir & Tishby, 2011), we know of no other spectral algorithm that optimizes the number of similarities needed.

## 3. Main Results

Before proceeding with our main results, we first clarify some notation and introduce a hierarchical clustering model that we will analyze. We refer to $\mathcal{A}$ as any flat clustering algorithm, which takes as parameters a dataset and a natural number $k$, indicating the number of clusters to produce. Throughout the paper, $k$ will denote the number of clusters at any split, and we will assume that $k$ is known and fixed across the hierarchy. We let $n$ be the number of objects in a datasets and define $s$ to be a parameter to our algorithms, influencing the number of measurements used by our algorithm, where smaller $s$ implies fewer measurements. The parameter $s$ reflects a tradeoff between the measurement overhead and the statistical accuracy of our algorithms; increasing $s$ increases the robustness of our method, albeit at the cost of requiring more measurements. Finally, our algorithms employ an abstract, possibly noisy similarity function $K$, which can model both cases where similarities are measured directly and where they are computed via some kernel function based on observed object features.

**Algorithm 1** ACTIVECLUSTER($\mathcal{A}, s, \{x_i\}_{i=1}^n, k$)
  **if** $n \leq s$ **then return** $\{x_i\}_{i=1}^n$
  Draw $S \subseteq \{x_i\}_{i=1}^n$ of size $s$ uniformly at random.
  $C'_1, \ldots C'_k \leftarrow \mathcal{A}(S, k)$.
  Set $C_1 \leftarrow C'_1, \ldots C_k \leftarrow C'_k$.
  **for** $x_i \in \{x_i\}_{i=1}^n \setminus S$ **do**
    $\forall j \in [k], \ \alpha_j \leftarrow \frac{1}{|C'_j|} \sum_{x_l \in C'_j} K(x_i, x_l)$.
    $C_{\arg\max_{j \in [k]} \alpha_j} \leftarrow C_{\arg\max_{j \in [k]} \alpha_j} \cup \{x_i\}$.
  **end for**
**output** $\{C_j, \text{ACTIVECLUSTER}(\mathcal{A}, s, C_j, k)\}_{j=1}^k$

**Definition 1** *A **hierarchical clustering** $\mathcal{C}$ on objects $\{x_i\}_{i=1}^n$ is a collection of clusters such that $C_0 \triangleq \{x_i\}_{i=1}^n \in \mathcal{C}$ and for each $C_i, C_j \in \mathcal{C}$ either $C_i \subset C_j, C_j \subset C_i$ or $C_i \cap C_j = \emptyset$. For any cluster $C$, if $\exists C'$ with $C' \subset C$, then there exists a set $\{C_i\}_{i=1}^k$ of disjoint clusters such that $\bigcup_{i=1}^k C_i = C$.*

Every hierarchical clustering $\mathcal{C}$ has a parameter $\eta$ that quantifies how balanced the clusters are at any split. Formally, $\eta \geq \max_{\text{splits}\{C_1,\ldots,C_k\}} \frac{\max_i |C_i|}{\min_i |C_i|}$, where each split is a non-terminal cluster, partitioned into $\{C_i\}_{i=1}^k$. $\eta$ upper bounds the ratio between the largest and smallest clusters sizes across all splits in $\mathcal{C}$. This type of balancedness parameter has been used in previous analyses of clustering algorithms (Eriksson et al., 2011; Balakrishnan et al., 2011), and it is common to assume that the clustering is not too unbalanced. For clarity of presentation, we will state our results assuming $\eta = O(1)$, although our proofs contain a precise dependence between the level of activeness $s$ and $\eta$.

### 3.1. An Active Clustering Framework

Our primary contribution is the introduction of a novel framework for hierarchical clustering that is efficient both in terms of the number of similarities used and the algorithmic running time. To recover any single split of the hierarchy, we run a flat clustering algorithm $\mathcal{A}$ on a small subset of the data to compute a seed clustering of the dataset. Using this initial clustering, we place each remaining object into the seed cluster for which it is most similar on average. This results in a flat clustering of the entire dataset, using only similarities to the objects in the small subset.

By recursively applying this procedure to each cluster, we obtain a hierarchical clustering, using a small fraction of the similarities. In this recursive phase, we do not observe any measurements between clusters at the previous split, i.e. to partition $C_j$, we only observe similarities between objects in $C_j$. This results in an *active* algorithm that focuses its measurements to resolve the higher-resolution cluster structure.

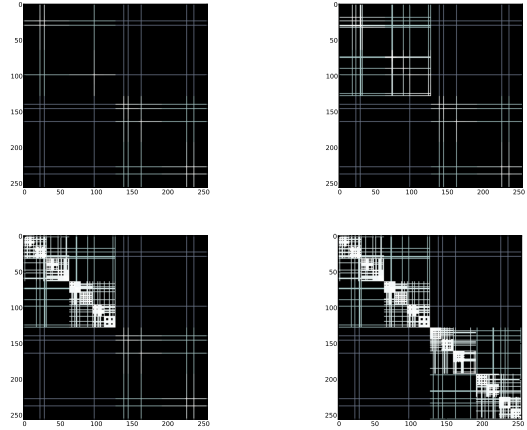

*Figure 1.* Sampling pattern of ACTIVECLUSTER.

Pseudocode for ACTIVECLUSTER is shown in Algorithm 1. As a demonstration, in Figure 1, we show the sampling pattern of ACTIVECLUSTER on the first, and second splits of a hierarchy (top row), in addition to the patterns at the end of the computation (bottom right). Only the similarities shown in white are needed. As is readily noticeable, the algorithm uses very few similarities yet can recover this hierarchy.

We are now ready to state our main theoretical contribution which characterizes ACTIVECLUSTER in terms of probability of success in recovering the true hierarchy (denoted $\mathcal{C}^\star$), measurement and runtime complexity. In order to make these guarantees we will need to place some mild restrictions on the similarity function $K$, which ensure that the similarities agree with the hierarchy (up to some random noise):

**K1** For each $x_i \in C_j \in \mathcal{C}^\star$ and $j' \neq j$:

$$\min_{x_k \in C_j} \mathbb{E}[K(x_i, x_k)] - \max_{x_k \in C_{j'}} \mathbb{E}[K(x_i, x_k)] \geq \gamma > 0$$

where expectations are taken with respect to the possible noise on $K$.

**K2** For each object $x_i \in C_j$, a set of $M_j$ objects of size $m_j$ drawn uniformly from cluster $C_j$ satisfies:

$$\mathbb{P}\left(\min_{x_k \in C_j} E[K(x_i, x_k)] - \sum_{x_k \in M_j} \frac{K(x_i, x_k)}{m_j} > \epsilon\right)$$
$$\leq 2\exp\{\frac{-2m_j \epsilon^2}{\sigma^2}\}$$

Where $\sigma^2 \geq 0$ parameterizes the noise on the similarity function $K$. Similarly, a set $M_{j'}$ of size $m_{j'}$ drawn uniformly from cluster $C_{j'}$ with $j' \neq j$ satisfies:

$$\mathbb{P}\left(\sum_{x_k \in M_{j'}} \frac{K(x_i, x_k)}{m_{j'}} - \max_{x_k \in C_{j'}} E[K(x_i, x_k)] > \epsilon\right)$$
$$\leq 2\exp\{\frac{-2m_{j'}\epsilon^2}{\sigma^2}\}$$

K1 states that the similarity from an object $x_i$ to its cluster should, in expectation, be larger than the the similarity from that object to any other cluster. This is related to the Tight-Clustering condition used in (Eriksson et al., 2011) and less stringent than earlier results which assume that within- and between-cluster similarities are constant and bounded in expectation (Rohe et al., 2010). Moreover, an assumption of this form seems necessary to ensure that even in expectation one could identify the clustering. Lastly, K2 enforces that within- and between-cluster similarities concentrate away from each other. This condition is satisfied, for example, if similarities are constant in expectation, perturbed with any subgaussian noise. We emphasize that K2 subsumes many of the assumptions of previous clustering analyses (for example (Balakrishnan et al., 2011; Rohe et al., 2010)).

**Theorem 1** *Let $\{x_i\}_{i=1}^n$ be a dataset with true hierarchical clustering $\mathcal{C}^\star$, let $K$ be a similarity function satisfing assumptions K1 and K2 and consider any flat clustering algorithm $\mathcal{A}$ with the following property:*

**A1** *For any dataset $\{y_i\}_{i=1}^m$ with clustering $\mathcal{C}'^\star$ where $K$ satisfies K1 and K2, $\mathcal{A}(\{y_i\}_{i=1}^m, k)$ returns the first split of $\mathcal{C}'^\star$ into $k$ clusters with probability $\geq 1 - o(\frac{k}{c_1 e^m})$ for some constant $c_1 > 0$.*

*Then* ACTIVECLUSTER$(\mathcal{A}, s, \{x_i\}_{i=1}^n, k)$:

**R1** *recovers all clusters of size at least $s$ with probability $1 - o(n^2 e^{-cs})$, for some constant $c = c(\eta, \gamma)$. This probability of success is $1 - o(1)$ as long as:*

$$s \geq \max\begin{Bmatrix} \frac{1}{c_1}\log n \\ 4(1+\eta)^2 \log n \\ 24\frac{1+\eta}{\gamma^2}\log(4C_\eta kn) \end{Bmatrix} = \Omega(\log(nk)) \quad (1)$$

**R2** *uses $O(ns\log n)$ similarity measurements.*
**R3** *runs in time $O(nA_s + ns\log n)$ where $\mathcal{A}$ on a datasets of size $s$ runs in time $O(A_s)$.*

At a high level, the theorem says that the clustering guarantee for a flat, non-active algorithm, $\mathcal{A}$, can be translated into a hierarchical clustering guarantee for an active version of $\mathcal{A}$, and that this active algorithm enjoys significantly reduced measurement and runtime complexity. The only property needed by $\mathcal{A}$ is that it recovers a flat clustering with very high probability. While the probability of success seems strangely high, we will show that for a fairly intuitive model, a simple spectral clustering algorithm enjoys this kind of guarantee. Verifying that the model satisfies the conditions K1 and K2, immediately results in a guarantee for the active version of this spectral algorithm.

Before delving into the proof of the theorem, some remarks are in order. First, by plugging in the lower bound for $s$ into the upper bound on the measurement complexity, we see that ACTIVECLUSTER needs $O(n\log(nk)\log n)$ similarities, which is considerably less than the $O(n^2)$ similarities required by a non-active algorithm. Second, at the lower bound for $s$, we see that unless $\mathcal{A}$ runs in exponential time, ACTIVECLUSTER runs in $\tilde{O}(n)$, which is significantly faster than *any* clustering algorithm that observes all of the similarities and must take $\Omega(n^2)$ time.

We now turn to the proof of R1. Due to space limitations, we defer many details and technical lemmas to the appendix. The proofs for R2 and R3 are straightforward, involving counting arguments on trees, and are also available in the appendix.

**Proof for R1:** We study the sampling, clustering and averaging phases of ACTIVECLUSTER in turn. In the sampling phase, we demonstrate that choosing $s$ objects at random does not result in a highly unbalanced subset. Using bernoulli concentration inequalities and a union bound we show that the balance factor across all splits is at most $2\eta + 1$ with probability $\geq 1 - o(ne^{-c_{\eta,\gamma}s})$, which goes to 1 under Equation 1.

For the clustering phase, Lemma 3 (in the appendix) shows that the total number of calls to $\mathcal{A}$ is at most $\frac{n}{k-1}$ and each time we call $\mathcal{A}$ we have a probability of success $\geq 1 - o(\frac{k}{e^{c_1 s}})$ by assumption A1. It is now easy to see that the probability of $\mathcal{A}$ failing at any split is $o(n\exp\{-s\})$, which is $o(1)$ under Equation 1.

In the averaging phase, we need to show that for each split of the hierarchy and object $x_i$, the sample average within cluster similarity is larger than the sample average between cluster similarity. Under assumption K1 and K2, we know that these quantities concentrate away from each other. Via a union bound across all objects and all levels of the hierarchy, we can conclude that the probability of making a mistake in any averaging procedure is $O(nk\log n \exp\{\frac{-\gamma^2 s}{4(1+\eta)}\})$ which again goes to zero as long as $s$ satisfies Equation 1.

**Algorithm 2** SPECTRALCLUSTER($W$)

Compute Laplacian $L = D - W$, $D_{ii} = \sum_{j=1}^n W_{ij}$
$v_2 \leftarrow$ smallest non-constant eigenvector of $L$.
$C_1 \leftarrow \{i : v_2(i) \geq 0\}$, $C_2 \leftarrow \{j : v_2(j) < 0\}$
**output** $\{C_1, C_2\}$.

### 3.2. Active Spectral Clustering

To make the guarantees in Theorem 1 more concrete, we show how to translate this into real guarantees for a specific subroutine algorithm $\mathcal{A}$. In this section, we turn a simple spectral algorithm (See pseudocode in Algorithm 2) into an active clustering algorithm, using the analysis from (Balakrishnan et al., 2011). The algorithm operates on hierarchically structured similarity matrices refered to as the **noisy Hierarchical Block Matrices** (again from (Balakrishnan et al., 2011)). These are defined as follows:

**Definition 2** *A similarity matrix $W$ is a* **noisy hierarchical block matrix** *(noisy HBM) if $W \triangleq A + R$ where $A$ is ideal and $R$ is a perturbation matrix:*

- *An* **ideal similarity matrix** *is characterized by ranges of off-block diagonal similarity values $[\alpha_\xi, \beta_\xi]$ for each cluster $C_\xi$ such that if $x \in C_{\xi \circ L}$ and $y \in C_{\xi \circ R}$, where $C_{\xi \circ L}$ and $C_{\xi \circ R}$ are two sub-clusters of $C_\xi$ at the next level in a binary hierarchy, then $\alpha_\xi \leq A_{x,y} \leq \beta_\xi$. Additionally, $\min\{\alpha_{\xi \circ R}, \alpha_{\xi \circ L}\} \geq \beta_\xi$.*
- *A symmetric $(n \times n)$ matrix $R$ is a* **perturbation matrix** *with parameter $\sigma$ if (a) $\mathbb{E}(R_{ij}) = 0$, (b) the entries of $R$ are subgaussian, that is $\mathbb{E}(\exp(tR_{ij})) \leq \exp\left(\frac{\sigma^2 t^2}{2}\right)$ and (c) for each row $i$, $R_{i1}, \ldots R_{in}$ are independent.*

To apply Theorem 1, we need to verify that the assumption K1 and K2 are met and SPECTRAL-CLUSTER succeeds with exponentially high probability. Checking that these conditions hold as long as $\sigma = O(1)$ results in the following guarantees for AC-TIVESPECTRAL, the active version of SPECTRALCLUSTER. Proof of this theorem is deferred to the appendix.

**Theorem 2** *Let $W$ be a noisy HBM with $\sigma = O(1)$ and $\eta = O(1)$. Then, ACTIVESPECTRAL succeeds in recovering all clusters of size $s$ with probability $1 - o(1)$ as long as Equation 1 holds. Moreover, ACTIVESPECTRAL uses $O(ns \log n)$ measurements and runs in $O(ns^2 \log s + ns \log n)$ time.*

The results of this theorem quantify one extreme of the tradeoff between statistical robustness and measurement complexity for hierarchical spectral algorithms. In particular, it states that ACTIVESPECTRAL can tolerate a constant amount of noise while using only $O(n \log^2 n)$ measurements. At the other end of this spectrum is the result of Balakrishnan et. al. (2011), showing that using $O(n^2)$ measurements, one can tolerate noise that grows fairly rapidly with $n$. Varying $s$ allows for interpolation between these two extremes.

### 3.3. Active $k$-means clustering

It is also possible to activize the popular $k$-means algorithm in our framework, but we cannot prove statistical performance guarantees since it is unknown whether $k$-means satisfies assumption A1. Activizing $k$-means helps illuminate the differences between observing similarities directly and computing similarities from directly observed object features. Conventionally, $k$-means fits into the latter framework. Here, the active version does not enjoy a reduced measurement complexity, because all of the objects must be observed, but it can reduce the number of similarity computations from $nkT$ to $skT + (n-s)kT$, since the iterative subroutine runs on only $s$ objects for $T$ iterations. In cases where the similarity function is expensive to compute, such as edit distance, this can lead to gains in running time.

A less traditional way to use $k$-means is to represent each object as a $n$-dimensional vector of its similarity to each other object. Here, we can apply $k$-means to a $n \times n$ similarity matrix, much like we can apply SPECTRALCLUSTER and this algorithm can be activized using our framework. While we cannot develop theoretical guarantees for this algorithm, which we call ACTIVEKMEANS, our experiments demonstrate that it performs very well in practice.

### 3.4. Some Practical Considerations

Our algorithm as stated has some shortcomings that enable theoretical analysis but that are undesirable for practical applications. Specifically, the fact that $k$ is known and constant across splits in the hierarchy, and the balancedness condition are both assumptions that are likely to be violated in any real-world setting. We therefore develop a variant of ACTIVESPECTRAL, called HEURSPEC, with several heuristics.

First, we employ the popular eigengap heuristic, in which the number of clusters $k$ is chosen so that the gap in eigenvalues $\lambda_{k+1} - \lambda_k$ of the Laplacian is large. Secondly, we propose discarding all subsampled objects with low degree (when restricted to the sample) in the hopes of removing underrepresented clusters from the sample. In the averaging phase, if an object is not highly similar to any cluster represented in the sample, we create a new cluster for this object. We expect that in tandem, these two heuristics will help us recover small clusters. By comparing the performance of HEURSPEC to that of ACTIVESPECTRAL, we indirectly evaluate these heuristics.

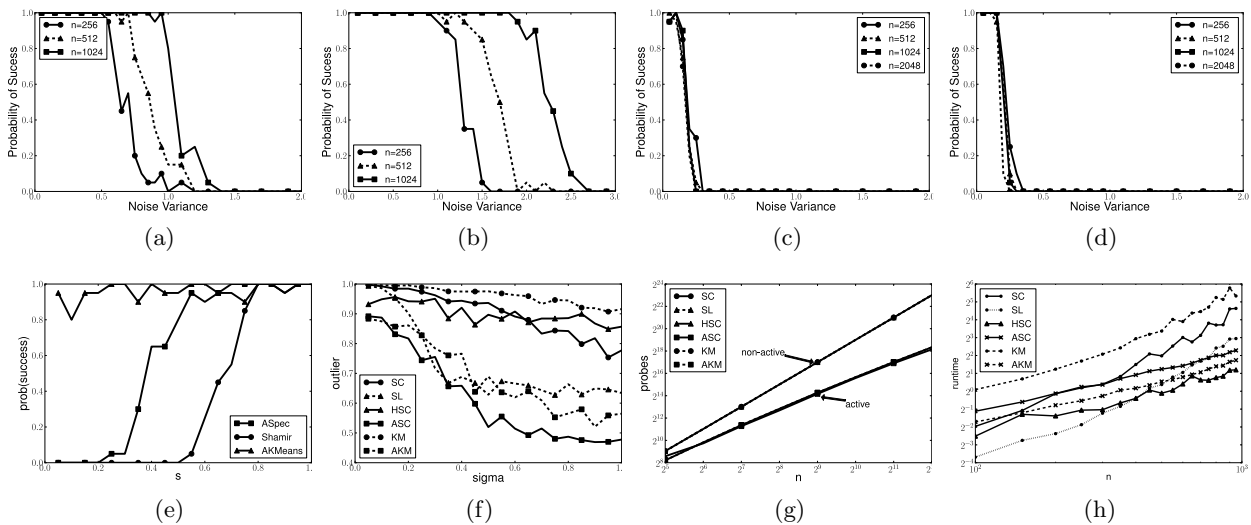

Figure 2. Simulation experiments. Top row: Noise thresholds for SPECTRALCLUSTER, K-MEANS, ACTIVESPECTRAL, and ACTIVEKMEANS with $s = \log(n)$ for active algorithms. Bottom row from left to right: probability of success as a function of $s$ for $n = 256, \sigma = 0.75$, outlier fractions on noisy HBM, probing complexity, and runtime complexity.

## 4. Simulations

In this section we present experiments that verify our theoretical results. By Theorem 2, we expect ACTIVESPECTRAL to be robust to a constant amount of noise $\sigma$, meaning that it will recover all sufficiently large splits with high probability. In comparison, Balakrishnan et. al. (2011), show that SPECTRALCLUSTER can tolerate noise growing with $n$. We constrast these guarantees by plotting the probability of successful recovery of the first split in a noisy HBM as a function of $\sigma$ for different $n$ in Figure 2. 2(a) demonstrates that indeed the noise tolerance of SPECTRALCLUSTER grows with $n$ while 2(c) demonstrates that ACTIVESPECTRAL enjoys constant noise tolerance. Figures 2(b) and 2(d) suggest that similar guarantees may hold for $k$-means and ACTIVEKMEANS.

Our theory also predicts that increasing the activeness parameter improves the statistical performance of ACTIVESPECTRAL. To demonstrate this, we plot the probability of successful recovery of the first split of a noisy HBM of size $n = 256$ as a function of $s$ for fixed noise variance. We compare three algorithms, ACTIVESPECTRAL, ACTIVEKMEANS, and Algorithm 1 from (Shamir & Tishby, 2011), which subsamples entries of the similarity matrix. In theory, ACTIVESPECTRAL requires $\Omega(n \log n)$ total measurements to recover a single split, whereas (Shamir & Tishby, 2011) show that their algorithm requires $\Omega(n \log^{3/2} n)$. Figure 2(e) demonstrates that this improvement is also noticeable in practice. ACTIVEKMEANS seems to enjoy an even more favorable dependence on $s$.

The simulations in Figures 2(a)-(e) only examine the ability of our algorithms to recover the first split of a hierarchy, while our theory predicts that all sufficiently large clusters can be reliably recovered. One way to measure this is the **outlier fraction** metric between the clustering returned by an algorithm and the true hierarchy (Eriksson et al., 2011). For any triplet of objects $x_i, x_j, x_k$ we say that the two clusterings **agree** on this triplet if they both group the same pair of objects deeper in the hierarchy relative to the third object and disagree otherwise. The outlier fraction is simply the fraction of triplets for which the two clusterings agree.

In Figure 2(f), we plot the outlier fraction for six algorithms as a function of $\sigma$ on the noisy HBM. The algorithms are: Hierarchical Spectral (SC), Single Linkage (SL), HEURSPEC (HSC), ACTIVESPECTRAL (ASC), Hierarchical $k$-Means (KM), and ACTIVEKMEANS (AKM). These experiments demonstrate that the non-active algorithms (except single linkage) are much more robust to noise than the corresponding active ones, as predicted by our theory, but also that the heuristics described in Section 3.4 have dramatic impact on performance.

Lastly, we verify the measurement and run time complexity guarantees for our active algorithms in comparison to the non-active versions. In Figure 2(g) and 2(h), we plot the number of measurements and running time as a function of $n$ on a log-log plot for each algorithm. The three non-active algorithms have steeper slopes than the active ones, suggesting that they are polynomially more expensive in both cases.

## 5. Real World Experiments

To demonstrate the practical performance of the ACTIVECLUSTER framework, we apply our algorithms to three real-world datasets and one additional synthetic dataset. The datasets are: The set of articles from NIPS volumes 0 through 12 from (Roweis, 2002), a subset of NPIC500 co-occurence data from the Read-the-Web project (Mitchell, 2009) which we call RTW, a SNP dataset from the HGDP (Pemberton et al., 2008), and a synthetic phylogeny dataset produced using `phyclust` (Chen, 2010). We refer the reader to the appendix for additional details on these datasets.

| Algorithm | HKM | HRC | Probes | Time (s) |
|---|---|---|---|---|
| SNP | | | | |
| HEURSPEC | 0.022 | 475 | 0.38 | 1350 |
| ACTIVESPEC | 0.019 | 19.1 | 0.13 | 450 |
| ACTIVEKMEANS | 0.018 | 12.5 | 0.12 | 420 |
| $k$-means | 0.0028 | 18.7 | 1 | 160 |
| Spectral | 0.0075 | 130 | 1 | 5660 |
| Phylo | | | | |
| HEURSPEC | 0.020 | 371 | 0.29 | 2500 |
| ACTIVESPEC | 0.012 | 22.9 | 0.071 | 600 |
| ACTIVEKMEANS | 0.012 | 25 | 0.071 | 555 |
| $k$-means | 0.0017 | 22.9 | 1 | 967 |
| Spectral | 0.0022 | 23.5 | 1 | 997 |
| NIPS | | | | |
| HEURSPEC | 0.0088 | 65.7 | 0.19 | 140 |
| ACTIVESPEC | 0.010 | 1.5 | 0.094 | 79.4 |
| ACTIVEKMEANS | 0.011 | 1.37 | 0.12 | 29 |
| $k$-means | 0.0017 | 1.66 | 1 | 723 |
| Spectral | 0.0033 | 6.30 | 1 | 26200 |
| RTW | | | | |
| HEURSPEC | 0.0079 | 18.1 | 0.41 | 419 |
| ACTIVESPEC | 0.0084 | 0.64 | 0.13 | 151 |
| ACTIVEKMEANS | 0.0073 | 0.485 | 0.22 | 70.9 |

Table 1. Real World Experiments

| Algorithm | SNP | Phylo |
|---|---|---|
| HEURSPEC | 0.596 | 0.878 |
| ACTIVESPEC | 0.374 | 0.971 |
| ACTIVEKMEANS | 0.383 | 0.94 |

Table 2. Outlier Fractions on Real Datasets

In the phylogeny and SNP datasets, we have access to a reference tree that can be used in our evaluation. In these cases we can report the outlier fraction, as we did in simulation. However, the other datasets lack such ground truth and without it, evaluating the performance of each algorithm is non-trivial. Indeed, there is no well-established metric for this sort of evaluation.

For this reason, we employ two distinct metrics to evaluate the quality of hierarchical clusterings. They are a hierarchical $K$-means objective (HKM) (Kauchak & Dasgupta, 2003) and an analogous hierarchical ratio-cut (HRC) objective, both of which are natural generalizations of the $k$-means and ratio cut objectives respectively, averaging across clusters, and removing small clusters as they bias the objectives. Formally, let $\mathcal{C}$ be the hierarchical clustering and let $\bar{\mathcal{C}}$ be all of the clusters in $\mathcal{C}$ that are larger than $\log n$. For each $C \in \bar{\mathcal{C}}$ let $x_C$ be the cluster center. Then:

$$\text{HKM}(\mathcal{C}) = \frac{1}{|\bar{\mathcal{C}}|} \sum_{C \in \bar{\mathcal{C}}} \frac{1}{|C|} \sum_{x_j \in C} \frac{x_j^T x_C}{||x_j|| ||x_C||} \quad (2)$$

$$\text{HRC}(\mathcal{C}) = \frac{1}{|\bar{\mathcal{C}}|} \sum_{C \in \bar{\mathcal{C}}} \sum_{C_k \subseteq C} \frac{K(C_k, C \setminus C_k)}{2|C_k|} \quad (3)$$

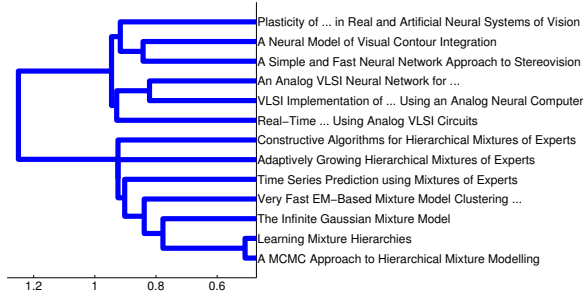

Figure 3. The ACTIVEKMEANS hierarchy restricted to a subset of NIPS articles.

In Table 1 and 2, we record experimental results across the datasets for our algorithms. On the read-the-web dataset, we were unable to run the non-active algorithms. On the SNP and phylogeny datasets, we include computing similarities via edit distance in the running time of each algorithm, noting that computing all pairs takes 6500 and 15000 seconds respectively. The immediate observation is that these algorithms are extremly fast; on the SNP and phylogeny datasets where computing similarities is the bottleneck, activization leads to significant performance improvements. Moreover, the algorithms perform well by our metrics; they find clusterings that score well according to HKM and HRC, or that have reasonable agreement with the reference clustering[1].

We are also interested in more qualitatively understanding the performance of these algorithms. For the NIPS data, we manually collected a small subset of articles and visualized the hierarchy produced by ACTIVEKMEANS restricted to these objects. The hierarchy in Figure 3 is what one would expect on the subset, attesting to the performance ACTIVEKMEANS. On the other hand, this same kind of evaluation on the RTW data demonstrates that active algorithms do not perform well on this dataset, while the non-active algorithms do. We suspect this is caused by the RTW dataset consisting of many small clusters that do not get sampled by the ACTIVECLUSTER framework.

---

[1] The SNP dataset is a $k$-way hierarchy and our algorithms (apart from HEURSPEC) recover binary hierarchies that cannot have high agreement with the reference.

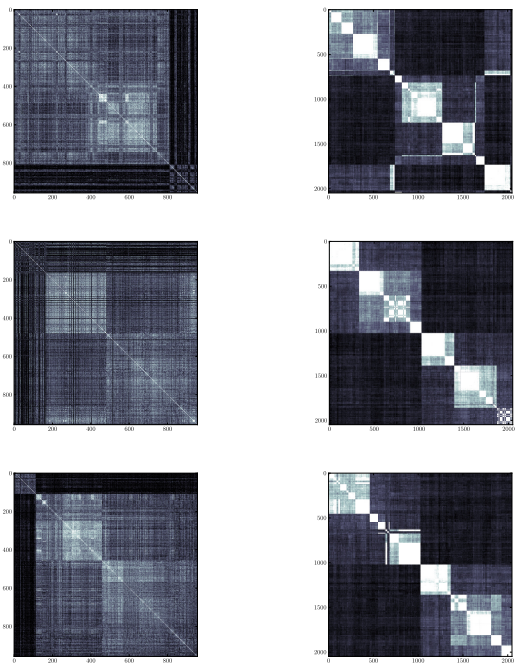

Figure 4. Heatmaps of permuted matrices for SNP (left) and Phylo (right). Algorithms are HEURSPEC, ACTIVESPECTRAL, and ACTIVEKMEANS from top to bottom.

For the SNP and phylogeny datasets, the permuted heatmaps are clear enough to be used in qualitative evaluations. These heatmaps are shown in Figure 4, and they suggest that all three active algorithms perform very well on these datasets. Heatmaps for the remaining datasets are less clear, but for completeness we include them in the appendix.

## 6. Discussion

Our results in this paper, showing that a family of active hierarchical clustering algorithms have strong performance guarantees, raise several interesting questions. We showed that ACTIVESPECTRAL enjoys reasonable statistical performance, but can other algorithms be activized while retaining statistical properties? Second, are there principled ways to circumvent a balancedness condition, even when objects are subsampled? Finally, is there a theoretically justified approach for estimating the number of clusters, $k$?

Another direction relates not toward clustering, but toward the recently popular matrix completion problem. On hierarchically structured matrices, our results imply that an active algorithm can recover high-rank (rank $n/\log n$) matrices using $O(n \log^2 n)$ similarities, an improvement over non-active approaches. Active algorithms may therefore yield impressive guarantees for matrix completion and related problems, and we hope to explore this direction in the future.

## Acknowledgements

This research is supported in part by AFOSR under grant FA9550-10-1-0382 and NSF under grant IIS-1116458. AK is supported in part by a NSF Graduate Research Fellowship.